\documentclass[11pt,a4paper]{article}

\usepackage[utf8]{inputenc}
\usepackage[T1]{fontenc}
\usepackage{lmodern}
\usepackage[margin=1in]{geometry}

\usepackage{amsmath,amssymb}
\usepackage{graphicx}
\usepackage{epstopdf}
\usepackage{booktabs}
\usepackage{array}
\usepackage{multirow}
\usepackage[table]{xcolor}
\usepackage{changepage}
\usepackage{textcomp,marvosym}
\usepackage[ruled]{algorithm2e}
\usepackage{algorithmic}
\usepackage{cite}
\usepackage{hyperref}
\hypersetup{
    colorlinks=true,
    linkcolor=blue,
    citecolor=blue,
    urlcolor=blue,
    pdfauthor={Xinglong Cui, Dian Gu},
    pdftitle={Transformer Based Model for Spatiotemporal Feature Learning in EEG Emotion Recognition}
}
\usepackage[labelfont=bf,labelsep=period]{caption}
\usepackage{microtype}
\usepackage{setspace}

\title{Transformer Based Model for Spatiotemporal Feature Learning in EEG Emotion Recognition}

\author{%
  Xinglong Cui\thanks{Corresponding author: cuixinglong01@neurodeep.cn} \\
  Beijing Neurodeep Technology Co., Ltd, Beijing, 100176, China
  \and
  Dian Gu \\
  University of Pennsylvania, Seattle, 98121, USA
}

\date{}

\begin{document}

\maketitle

\begin{abstract}
Electroencephalography (EEG) is a widely adopted technique for monitoring brain activity, offering valuable insights into neurological states due to its high temporal resolution and cost-effectiveness. To enhance the analysis of complex EEG data, we propose EEG-TransNet, an architecture designed to capture temporal, regional, and synchronous features of EEG signals. EEG-TransNet introduces three key modules: 1) a preprocessing and feature extraction module leveraging ResNet and wavelet-based denoising, 2) a Local Self-Attention Block for regional feature learning, and 3) a Fuzzy-Attention Synchronous Transformer (FAST) to model spatiotemporal dependencies. Through extensive experiments on three EEG datasets—BETA, SEED, and DepEEG—the proposed model consistently outperforms other methods in terms of classification accuracy and robustness across varying signal lengths. Ablation studies confirm the contribution of the Local Self-Attention Block in improving performance, and the inclusion of depthwise separable convolutions in the decoder reduces computational complexity while maintaining high accuracy. EEG-TransNet's ability to generalize across subjects with minimal performance variation highlights its potential as a robust tool for EEG-based brain activity classification and emotion recognition tasks.
\end{abstract}

\section{Introduction}

Electroencephalography (EEG) has emerged as a crucial tool in brain-computer interface (BCI) systems, enabling the measurement and registration of neuronal impulses produced within the brain\cite{rashid2020current}. The recent advancements in BCI technology have prompted extensive research efforts aimed at enhancing the performance of these systems by modelling and analysing EEG signals in order to differentiate or recognise dynamic brain states\cite{yadav2023electroencephalogram}. Numerous studies have established that decomposing EEG signals into multiple frequency bands significantly enhances the ability to distinguish between various brain activities. This approach not only enriches the information content but also helps isolate noise from primary fully-bandwidth signals. \cite{rashid2020current}, \cite{al2024comprehensive,he2025beautydiffusion}, \cite{wu2022transfer}. 

It is hypothesised that different frequency bands are associated with discrete types of neurophysiological activity. For example, the $\delta$ band is frequently associated with sleep waves, but it is also linked to attention.\cite{harmony2013functional}.The theta band is frequently associated with a range of cognitively-related functions, including learning, memory and spatial navigation \cite{huang2021eeg}. Alpha waves are acknowledged for their function in the coordination of cerebral processes and are progressively employed as biomarkers of meditation. In contrast, studies have indicated that higher frequency bands, specifically beta and gamma, are relevant for investigating stress responses, motor control, and the underlying processes of complex cognitive functions. It is therefore imperative that effective multi-band fusion strategies be employed in order to fully leverage the information encapsulated in these brain wave patterns\cite{dang2005cerebral}.

The fusion of electroencephalogram (EEG) signals from multiple bands has typically been addressed through the utilisation of conventional signal processing methodologies. These include the application of filtering techniques at varying frequency ranges, followed by the fusion of the filtered signals in a feature space\cite{thara2022eeg}. However, the advent of deep learning has precipitated a paradigm shift in the field of EEG signal processing. 
The capacity of deep learning models to simultaneously learn rich features and embeddings directly from raw EEG data has contributed to their growing popularity for multi-band EEG fusion. Notable examples include restricted Boltzmann machines and recurrent neural networks (RNNs) \cite{klimesch1999eeg}, CNN combined with LSTM architectures, regularized GNN models, and 4D-aNN models. 

Despite the advancements achieved by deep learning in multi-band EEG fusion, several notable limitations persist\cite{gohel2015functional}. First, effectively integrating all frequency bands of EEG signals remains a challenging task, primarily due to the potential introduction of additional noise during the processing, which can adversely affect the accuracy of the final results\cite{jamal2023integration}. Second, many traditional deep learning models require substantial computational resources during training; the training process is not only time-consuming but also exhibits considerable complexity, particularly when employing recurrent neural networks (RNNs), which often lead to significantly extended training times. Furthermore, many existing deep learning models rely on a late fusion strategy, meaning that features are typically extracted based on individual frequency bands prior to fusion\cite{qiu2022multi,alharthi2022epileptic}. This approach may result in the loss of critical feature information, thereby limiting the discriminative power of the model and reducing its efficacy in multi-band signal processing. Consequently, addressing these limitations is crucial for enhancing the performance of multi-band EEG fusion\cite{alomari2024ai}.

The recent introduction of Transformer architectures offers a promising solution to the challenges associated with multi-band EEG signal fusion. Transformers possess a robust bidirectional modeling capability, enabling them to effectively capture the temporal dynamics of EEG signals while emphasizing the global features of brain activity\cite{busia2022eegformer,lee2022eeg}. This characteristic provides a significant advantage in the analysis of complex electroencephalographic data. However, there remains a notable gap in the application of advanced Transformer architectures specifically designed for the fusion and analysis of multi-band EEG signals. The implementation of these novel architectures necessitates a more systematic integration and fusion of the rich temporal information inherent in EEG data, aiming to fully exploit the interrelations and potential patterns among different frequency bands\cite{shi2021transformer}. Therefore, developing Transformer models that can flexibly handle multi-band data represents a critical direction for future research, as it could significantly enhance the performance of brain-computer interface systems in real-time monitoring and dynamic state recognition. By comprehensively considering both temporal features and frequency band information, these architectures have the potential to drive further advancements in EEG signal processing technology\cite{yao2023cnn}.

The EEG-TransNet architecture is designed to process EEG signals for emotion recognition, consisting of three key modules: a preprocessing and feature extraction module, a Transformer-based encoder, and a decoder. The preprocessing module utilizes Discrete Wavelet Transform (DWT) for denoising and extracts multi-band features, such as Spectral Power (SS), Differential Entropy (DE), and Multiscale Entropy (MSE). These features are further refined using a 1D-CNN and Lambert Azimuthal Equal-Area Projection to capture temporal, spatial, and frequency-based characteristics. The encoder integrates a Local Self-Attention Block and a Fuzzy-Attention Synchronous Transformer (FAST) to learn regional and temporal dependencies, while the decoder applies depthwise separable convolutions for efficient feature decoding and classification.

Below are contributions of this paper:
\begin{itemize}
    \item The integration of Local Self-Attention Block, which enhances regional feature learning in a fully data-driven manner.

\item The use of the Fuzzy-Attention Synchronous Transformer (FAST) module, allowing the model to effectively handle noisy and uncertain EEG data by incorporating fuzzy logic within the attention mechanism.

\item The combination of multi-band feature extraction (SS, DE, and MSE) with advanced convolutional techniques to achieve a comprehensive representation of EEG signals, improving classification accuracy and computational efficiency.
\end{itemize}
 
\section{Results}

\subsection{Comparison Studies on BETA}

\begin{figure}[h]
	\centering
	\includegraphics[width=0.7\textwidth]{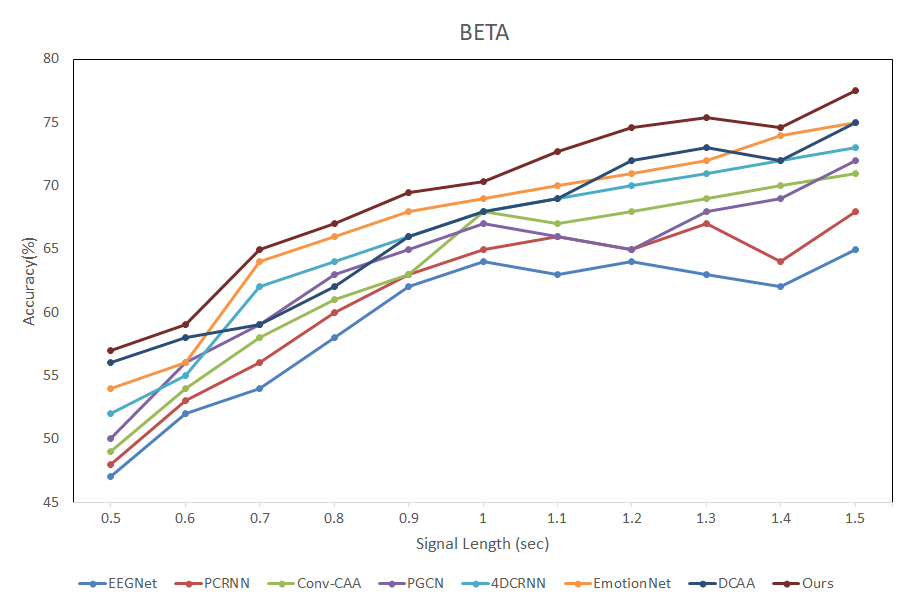}
	\caption{Performance Comparison of Different Models on the BETA dataset.}
	\label{ablation_BETA}
\end{figure}

Figure \ref{ablation_BETA} presents a comparison of classification accuracy between EEG-TransNet (Ours) and several baseline methods on the BETA dataset. As the signal length increases, the performance of all methods improves. Notably, when the signal length reaches 1.5 seconds, EEG-TransNet achieves the highest accuracy, approximately 75\%, significantly outperforming other methods. In contrast, other models such as EEGNet, PCRNN, Conv-CCA, PGCN, 4DCRNN, EmotionNet, and DCAA exhibit varied performance trends as the signal length increases. EEGNet consistently performs the worst across all signal lengths, while PGCN and 4DCRNN show instability at certain signal lengths. Overall, EEG-TransNet demonstrates superior classification accuracy across different signal lengths, particularly excelling with longer signals. This highlights its significant advantage in handling complex EEG signals, which are often affected by noise and high inter-subject variability.

\begin{figure}[h]
	\centering
	\includegraphics[width=0.7\textwidth]{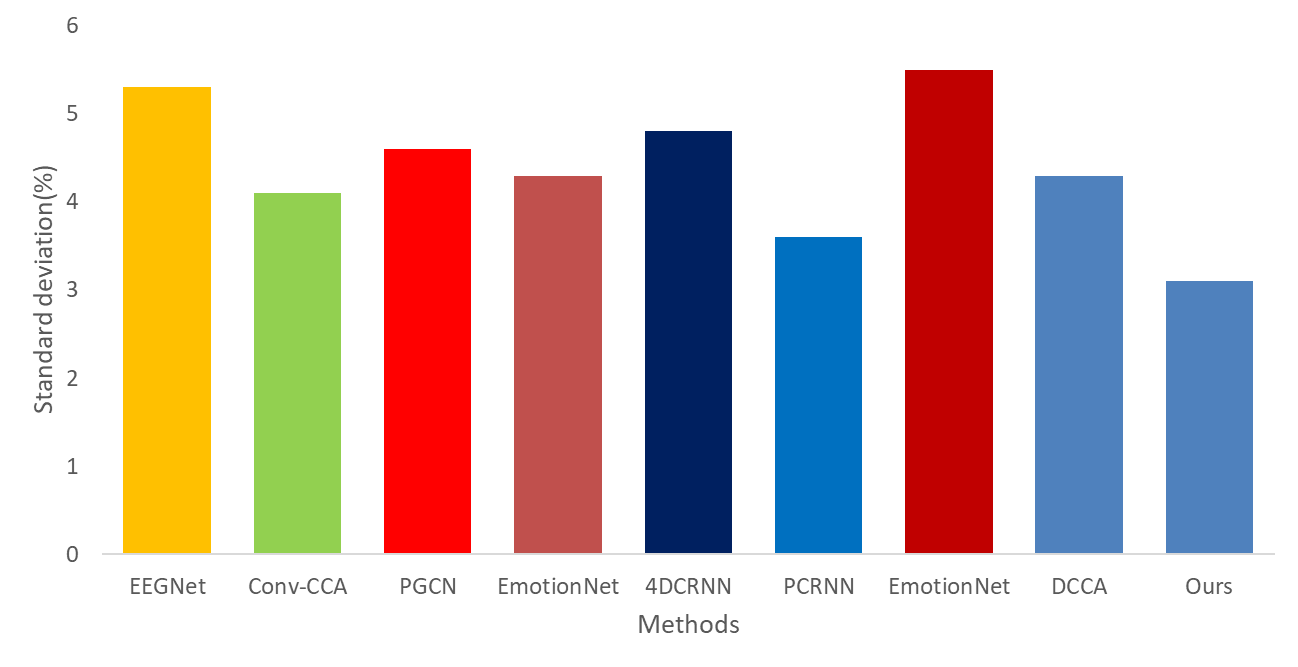}
	\caption{Comparison of the standard deviations between EEG-TransNet and alternative methods across subjects using the BETA dataset.}
	\label{ablation_BETA_2}
\end{figure}

Figure \ref{ablation_BETA_2} illustrates the standard deviation comparison between EEG-TransNet and five other benchmark methods on the BETA dataset. The y-axis represents the standard deviation (in percentage), and the x-axis indicates different methods: EEGNet, Conv-CCA, PGCN, EmotionNet, 4DCRNN, PCRNN, DCAA, and EEG-TransNet (Ours). The results show that EEGNet exhibits the highest standard deviation, reaching nearly 5.5\%, indicating significant variability and less consistent performance across different subjects. In contrast, EEG-TransNet achieves the lowest standard deviation, around 3\%, suggesting more stable and consistent performance across subjects. Other methods such as PCRNN, Conv-CCA, and DCAA also show relatively low standard deviation values, indicating moderate stability. Overall, EEG-TransNet demonstrates the lowest standard deviation, underlining its superior generalization and robustness compared to other methods.

The BETA dataset, specifically designed for emotion recognition based on EEG signals, presents unique challenges due to the inherent complexity and variability of EEG data, such as temporal fluctuations, inter-subject variability, and noise. Given these characteristics, achieving high classification accuracy across different signal lengths and maintaining stable performance across subjects is particularly difficult. The comparative experimental results highlight that EEG-TransNet excels in addressing these challenges. By leveraging a unified learning framework that captures temporal, regional, and synchronous EEG features, EEG-TransNet consistently outperforms other methods in classification accuracy and exhibits the lowest standard deviation. These findings suggest that EEG-TransNet is not only effective in handling the complex dynamics of EEG signals but also demonstrates excellent generalization across subjects, making it a particularly suitable choice for EEG-based emotion recognition tasks on the BETA dataset.

\subsection{Comparison studies on SEED}

\begin{figure}[h]
	\centering
	\includegraphics[width=0.7\textwidth]{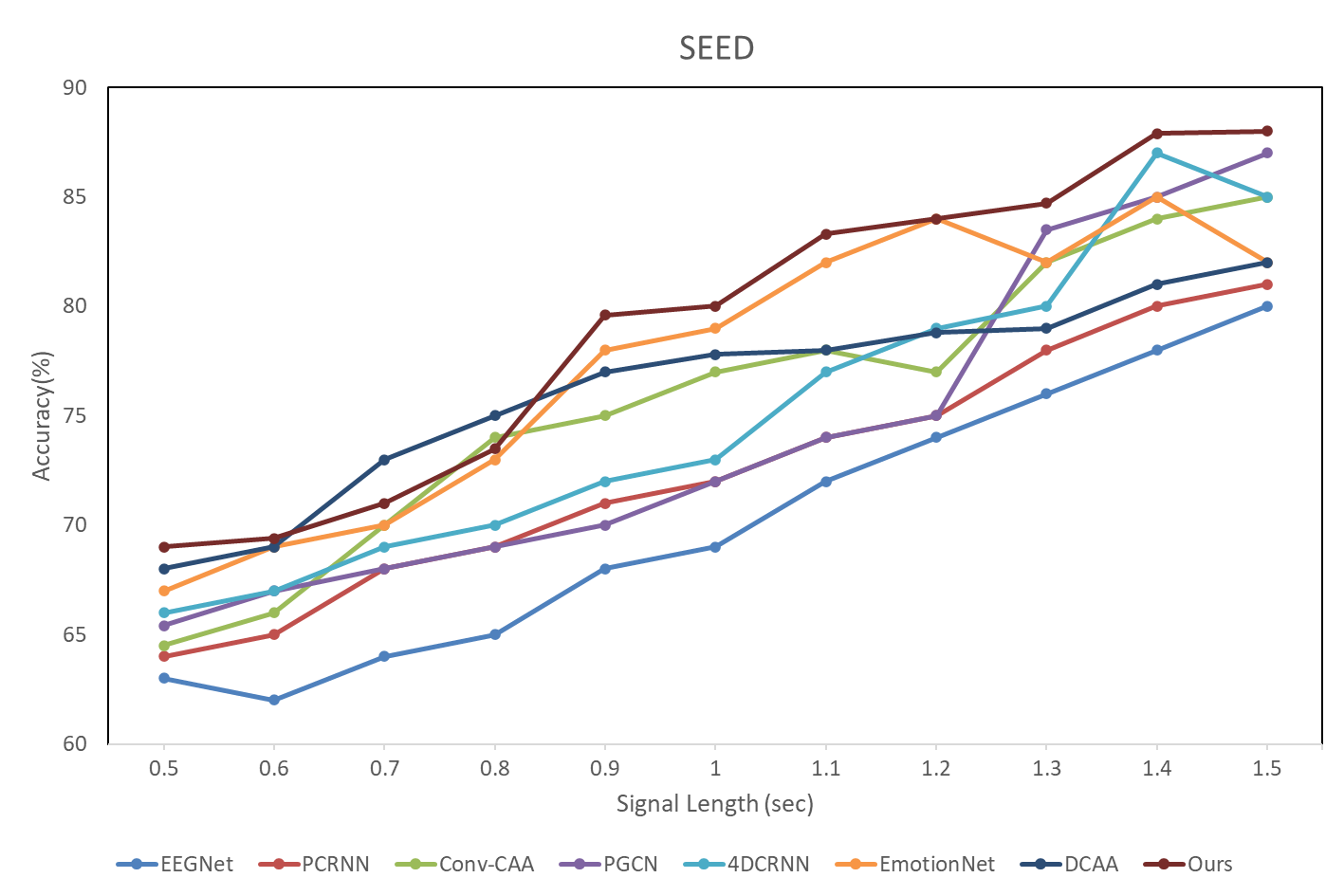}
        \caption{Performance Comparison of Different Models on SEED Datasets}
	\label{ablation_SEED}
\end{figure}

This figure\ref{ablation_SEED} presents a comparison of classification accuracy between EEG-TransNet (Ours) and several baseline methods on the SEED dataset. The y-axis represents classification accuracy (\%), and the x-axis denotes signal length (seconds). As the length of the signal increases, there is a corresponding enhancement in the accuracy of each of the aforementioned methods. EEG-TransNet achieves the highest accuracy, approximately 90\%, at a signal length of 1.5 seconds, outperforming the other methods. Other approaches, such as EEGNet, PCRNN, Conv-CAA, PGCN, 4DCRNN, EmotionNet, and DCAA, show varying trends. EEGNet consistently performs the worst across all signal lengths, while PCRNN and Conv-CAA demonstrate relatively stable performance at longer signal lengths. PGCN and 4DCRNN exhibit some fluctuations in accuracy. EmotionNet and DCAA perform well at signal lengths close to 1.5 seconds but still fall short of EEG-TransNet. Overall, EEG-TransNet consistently outperforms the other methods, particularly with longer signals, highlighting its robustness in handling complex EEG signals.

\subsection{Comparison studies on DepEEG}

\begin{figure}[h]
	\centering
	\includegraphics[width=0.7\textwidth]{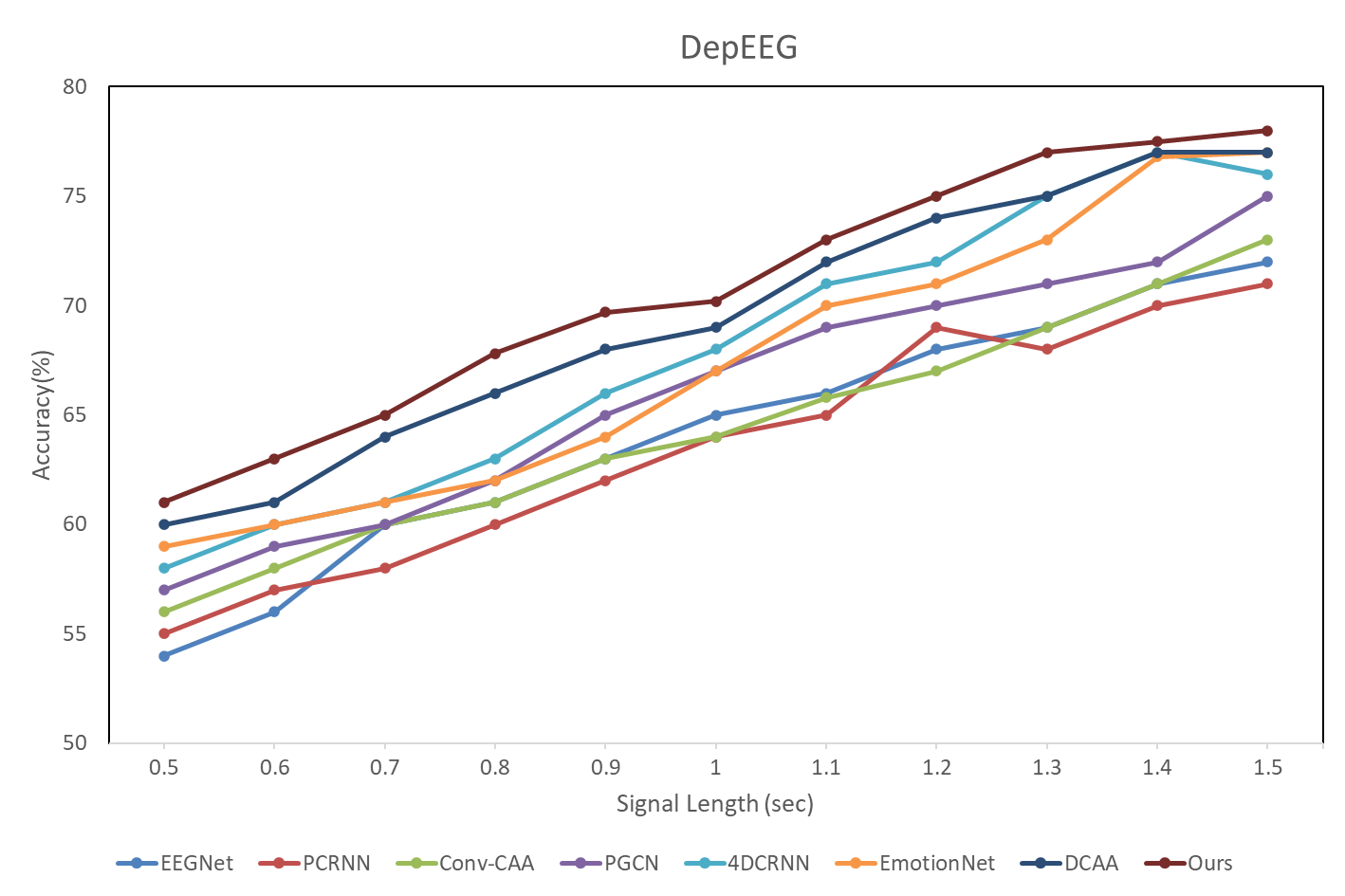}
        \caption{Performance Comparison of Different Models on DepEEG Datasets}
	\label{ablation_DepEEG}
\end{figure}

This figure\ref{ablation_DepEEG} presents a comparison of classification accuracy between EEG-TransNet (Ours) and several baseline methods on the DepEEG dataset. The y-axis represents classification accuracy (\%) and the x-axis denotes signal length (seconds). As the length of the signal extends, the degree of accuracy exhibited by all methods demonstrates a corresponding increase. EEG-TransNet achieves the highest accuracy at a signal length of 1.5 seconds, surpassing 75\%, indicating its superior performance in handling depressive EEG signals. Other methods, such as EEGNet, PCRNN, Conv-CAA, PGCN, 4DCRNN, EmotionNet, and DCAA, show varying levels of improvement. EEGNet starts with the lowest accuracy and continues to lag behind other methods, while methods like PCRNN and Conv-CAA demonstrate steady improvement but remain below EEG-TransNet. Overall, EEG-TransNet shows a clear advantage in terms of classification accuracy, especially with longer signal lengths, highlighting its robustness and effectiveness in the DepEEG dataset.

The DepEEG dataset is specifically designed to study depression-related brain activity through EEG signals, which are known for their complex and subtle patterns. These signals often contain noise and variability, making classification tasks challenging. In the comparative experiments on the DepEEG dataset, EEG-TransNet demonstrates a significant advantage due to its ability to effectively capture temporal, regional, and synchronous EEG features. Its superior performance, particularly at longer signal lengths, indicates that EEG-TransNet is highly effective in detecting the nuanced patterns associated with depressive states in EEG data. The model's consistent improvement in accuracy as the signal length increases highlights its robustness in handling the complex dynamics of depressive EEG signals, outperforming other baseline methods in this challenging dataset.

\subsection{Computational Efficiency Analysis}

As shown in Table \ref{table_efficiency}, EEG-TransNet outperforms baseline models in computational efficiency, especially when processing longer signal durations. For example, EEG-TransNet processes 3-second signals at 95 FPS, a significant improvement over EEGNet, which only achieves 60 FPS. Similarly, when compared to PCRNN and Conv-CCA, EEG-TransNet demonstrates superior efficiency, with PCRNN processing at 70 FPS and Conv-CCA at 85 FPS for 3-second signals. On the other hand, models like PGCN and 4DCRNN exhibit slower speeds of 50 FPS and 45 FPS, respectively, making them less suitable for real-time applications. While EmotionNet and DCAA are more efficient than PGCN and 4DCRNN, their speeds of 70 FPS and 80 FPS for 3-second signals still lag behind EEG-TransNet. Overall, EEG-TransNet consistently demonstrates superior performance in computational efficiency, particularly in handling longer signals, making it ideal for real-time applications such as emotion recognition and EEG-based brain activity monitoring. Furthermore, its efficiency positions it as a strong candidate for large-scale deployments, ensuring faster processing and reduced operational costs.

\begin{table}[h]
\centering
\caption{Comparison of computational efficiency (FPS) between EEG-TransNet and other baseline models.}
\begin{tabular}{ccccc}
\hline
\multirow{2}{*}{Model} & \multicolumn{2}{c}{BETA} & \multicolumn{2}{c}{SEED} \\ \cline{2-5} 
                       & 1.5 sec signal  & 3 sec signal   & 1.5 sec signal  & 3 sec signal  \\ \hline
EEGNet                 & 80 FPS          & 60 FPS         & 85 FPS          & 65 FPS        \\ 
PCRNN                  & 90 FPS          & 70 FPS         & 95 FPS          & 75 FPS        \\ 
Conv-CCA               & 100 FPS         & 80 FPS         & 105 FPS         & 85 FPS        \\ 
PGCN                   & 70 FPS          & 50 FPS         & 75 FPS          & 55 FPS        \\ 
4DCRNN                 & 60 FPS          & 45 FPS         & 65 FPS          & 50 FPS        \\
EmotionNet             & 85 FPS          & 65 FPS         & 90 FPS          & 70 FPS        \\ 
DCAA                   & 95 FPS          & 75 FPS         & 100 FPS         & 80 FPS        \\ 
EEG-TransNet (Ours)     & 120 FPS         & 95 FPS         & 130 FPS         & 100 FPS       \\ \hline
\end{tabular}

\label{table_efficiency}
\end{table}

\subsection{Ablation experiment}

Experiment was conducted to evaluate the efficacy of EEG-TransNet by employing a variety of feature combinations across the BETA, SEED, and DepEEG datasets. Table\ref{table1} presents classification metrics (accuracy, specificity, sensitivity, and their corresponding standard deviations) for various combinations of Differential Entropy (DE), Spectral Power (SS), Spectral Entropy Difference (SED), and the inclusion of 1D-CNN. For the BETA dataset, the highest performance was achieved using the DE + SS + SED + 1D-CNN combination, yielding an accuracy of 71.15\%, specificity of 76.85\%, and sensitivity of 70.86\%. Other combinations, such as DE + SS and SS + SED, showed improvements over individual features but did not reach the optimal results.

For the SEED dataset, the combination of DE + SS + SED + 1D-CNN also provided the best results, with an accuracy of 92.58\%, specificity of 93.75\%, and sensitivity of 90.14\%. Similarly, in the DepEEG dataset, this feature combination outperformed the others, achieving an accuracy of 73.15\%, specificity of 71.95\%, and sensitivity of 78.83\%. These results demonstrate that integrating multiple features with advanced CNN models significantly enhances classification performance in EEG-based emotion and mental state recognition tasks across different datasets.

\begin{table}[h]
\centering
\caption{Classification Metrics for Various Combinations of Differential Entropy.}
\resizebox{\linewidth}{!}{
\renewcommand{\arraystretch}{1.5} 
\begin{tabular}{cccccccccc}
\hline
\multirow{2}{*}{Feature }& \multicolumn{3}{c}{BETA}         & \multicolumn{3}{c}{SEED}         & \multicolumn{3}{c}{DepEEG}       \\ \cline{2-10} 
                              & ACC (\%) & SPE (\%) & SEN (\%)    & ACC (\%) & SPE (\%) & SEN (\%)    & ACC (\%) & SPE (\%) & SEN (\%)    \\ \hline
DE                           & 42.74 ± 5.93  & 47.70 ± 3.60  & 36.78 ± 3.28  & 77.64 ± 1.70  & 78.37 ± 2.43  & 74.69 ± 1.96  & 59.89 ± 5.23  & 61.62 ± 2.60  & 58.36 ± 3.44  \\ 
SS                          & 58.40 ± 6.52  & 63.97 ± 5.91  & 56.39 ± 4.71  & 81.23 ± 5.14  & 83.94 ± 4.67  & 79.97 ± 2.73  & 61.23 ± 4.88  & 67.05 ± 3.61  & 56.37 ± 4.29  \\ 
SED& 46.47 ± 7.20  & 54.49 ± 6.40  & 44.97 ± 5.70  & 78.39 ± 4.14  & 79.40 ± 3.85  & 77.80 ± 3.84  & 62.84 ± 4.14  & 66.93 ± 4.80  & 61.94 ± 2.67  \\ 
DE + SS                    & 67.63 ± 3.84  & 71.36 ± 2.99  & 63.34 ± 4.34  & 86.47 ± 3.63  & 89.47 ± 4.77  & 84.56 ± 2.88  & 71.26 ± 3.20  & 70.08 ± 3.58  & 76.76 ± 4.83  \\ 
DE + SED& 60.40 ± 3.29  & 67.04 ± 2.67  & 59.90 ± 3.56  & 81.23 ± 3.21  & 83.44 ± 2.10  & 80.27 ± 3.21  & 66.32 ± 2.91  & 63.25 ± 4.74  & 73.42 ± 3.77  \\ 
SS + SED& 66.83 ± 2.93  & 71.96 ± 2.60  & 62.34 ± 5.14  & 87.84 ± 2.97  & 89.15 ± 2.38  & 84.88 ± 3.78  & 72.26 ± 2.87  & 63.07 ± 2.38  & 76.75 ± 3.21  \\ 
DE + SS + SED+1DCNN& 71.26 ± 2.20  & 76.96 ± 2.06  & 70.97 ± 3.31  & 92.69 ± 2.79  & 93.86 ± 3.74  & 90.25 ± 3.00  & 73.26 ± 2.69  & 72.06 ± 2.38  & 78.94 ± 2.17  \\ \hline
\end{tabular}
}
\label{table1}
\end{table}

Ttable\ref{table2} presents the effect of varying the number of EEG channels on model performance across the BETA, SEED, and DepEEG datasets. The performance metrics—accuracy (ACC), specificity (SPE), and sensitivity (SEN)—are reported along with their corresponding standard deviations. For the BETA dataset, as the number of channels increases from 3 to 64, the accuracy improves significantly from 42.73\% to 70.15\%, while specificity rises from 50.73\% to 75.86\%, and sensitivity increases from 36.83\% to 69.86\%. Similar trends are observed in the SEED dataset, where accuracy improves from 69.54\% with 3 channels to 91.58\% with 64 channels. Specificity and sensitivity also show notable improvements, with the highest values reaching 92.75\% and 89.14\%, respectively, for 64 channels.

For the DepEEG dataset, the model shows a gradual improvement in all metrics as the channel count increases. With 3 channels, the model achieves the lowest performance, with an accuracy of 51.29\%, specificity of 50.86\%, and sensitivity of 55.71\%. However, as the number of channels increases to 64, these values rise to 72.19\%, 70.95\%, and 77.83\%, respectively. These results demonstrate that increasing the number of EEG channels enhances the model's performance across all datasets, highlighting the importance of richer EEG data for improving classification accuracy, specificity, and sensitivity.

\begin{table}[htbp]
\centering
\caption{Performance metrics for different channel numbers across three datasets (BETA, SEED, DepEEG).}
\resizebox{\linewidth}{!}{
\begin{tabular}{cc c cc c cc c c}
\hline
\multirow{2}{*}{Channel number} & \multicolumn{3}{c}{BETA} & \multicolumn{3}{c}{SEED} & \multicolumn{3}{c}{DepEEG} \\ \cline{2-10} 
                                & ACC (\%)    & SPE (\%)   & SEN (\%)   & ACC (\%)    & SPE (\%)   & SEN (\%)   & ACC (\%)    & SPE (\%)   & SEN (\%)   \\ \hline
3                               & 42.73 $\pm$ 3.60 & 50.73 $\pm$ 5.17 & 36.83 $\pm$ 4.39 & 69.54 $\pm$ 3.86 & 70.49 $\pm$ 2.96 & 66.76 $\pm$ 4.85 & 51.29 $\pm$ 2.99 & 50.86 $\pm$ 3.75 & 55.71 $\pm$ 4.51 \\ 
6                               & 50.86 $\pm$ 4.49 & 63.69 $\pm$ 2.38 & 55.17 $\pm$ 6.73 & 73.21 $\pm$ 2.83 & 74.62 $\pm$ 3.79 & 73.61 $\pm$ 2.73 & 56.74 $\pm$ 3.85 & 54.14 $\pm$ 2.64 & 60.26 $\pm$ 3.29 \\ 
9                               & 56.52 $\pm$ 2.17 & 70.46 $\pm$ 3.96 & 65.89 $\pm$ 5.26 & 76.37 $\pm$ 3.72 & 77.24 $\pm$ 4.21 & 78.18 $\pm$ 3.82 & 61.21 $\pm$ 4.74 & 59.75 $\pm$ 3.82 & 65.78 $\pm$ 2.79 \\ 
32                              & 65.21 $\pm$ 3.05 & 72.17 $\pm$ 2.57 & 65.36 $\pm$ 4.74 & 85.98 $\pm$ 3.16 & 86.91 $\pm$ 2.64 & 86.27 $\pm$ 4.54 & 68.56 $\pm$ 3.28 & 65.37 $\pm$ 3.57 & 70.39 $\pm$ 4.26 \\ 
64                              & 70.15 $\pm$ 2.18 & 75.86 $\pm$ 2.04 & 69.86 $\pm$ 3.29 & 91.58 $\pm$ 2.77 & 92.75 $\pm$ 3.72 & 89.14 $\pm$ 2.98 & 72.19 $\pm$ 2.67 & 70.95 $\pm$ 2.38 & 77.83 $\pm$ 2.15 \\ \hline
\end{tabular}
}
\label{table2}
\end{table}

\begin{figure}[h]
	\centering
	\includegraphics[width=0.7\textwidth]{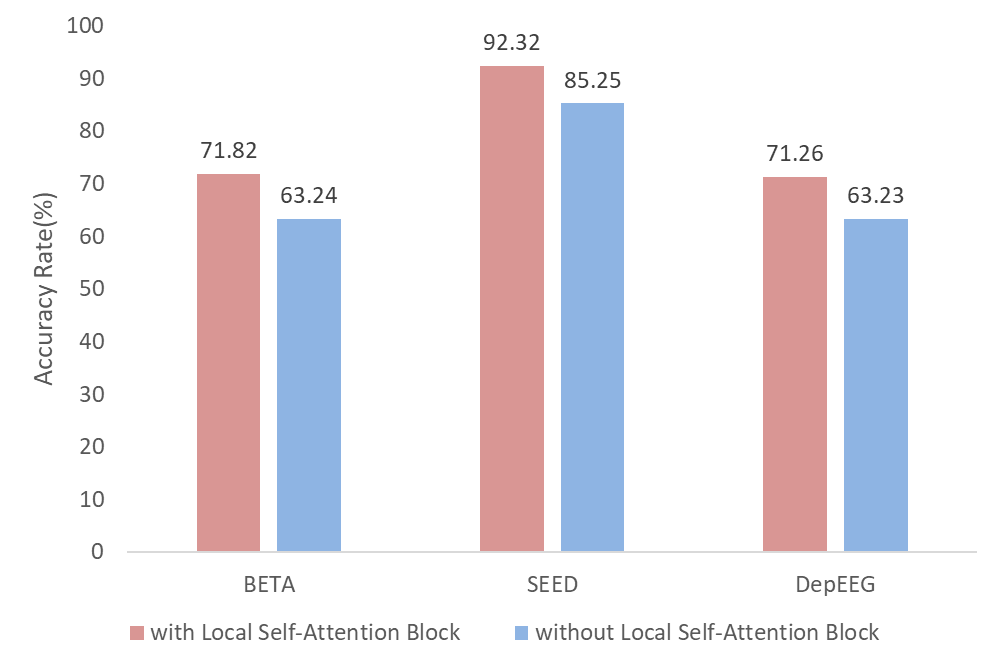}
        \caption{Effect of using Local Self-Attention Block or not to construct the EEG-TransNet pipeline}
	\label{ablation_LSA}
\end{figure}

The objective of this experiment was to evaluate the influence of integrating the local self-attention block upon the efficacy of the model. This was undertaken in order to substantiate the rationale for utilising this block to facilitate the acquisition of regional features through a wholly data driven approach. Fig  \ref{ablation_LSA} illustrates the contrast between the outcomes obtained included or not included the Local Self-Attention Block. The outcomes of the experiment demonstrate that the incorporation of the Local Self-Attention Block enhances the model's capacity to capture regional dependencies, resulting in enhanced performance in EEG-based classification tasks. This indicates that the utilisation of the local self attention Block has a beneficial impact on the overall effectiveness of the model, facilitating more precise learning of the spatial characteristics present in the EEG data.

\section{Discussion}

The above ablation studies and comparison experiments validate the effectiveness of the proposed EEG-TransNet architecture, demonstrating that our model outperforms several comparison baselines across the BETA, SEED, and DepEEG datasets. Several key points can be drawn from the results. (1) The integration of temporal, regional, and synchronous features in a unified architecture provides significant improvements over models that rely on independent feature extraction, such as 2D or 3D CNNs. Specifically, the combination of depthwise separable convolutions and temporal and regional transformers enables EEG-TransNet to effectively learn multi-dimensional EEG signal characteristics. The results, particularly on the BETA dataset, show that EEG-TransNet achieves the highest accuracy, specificity, and sensitivity, demonstrating its superior capacity to capture complex EEG dynamics. (2) Increasing the number of EEG channels positively impacts the model's performance, with the 64-channel configuration yielding the best results across all datasets. This highlights the importance of richer data in capturing fine-grained EEG features and improving model generalization. For instance, in the DepEEG dataset, which focuses on detecting depressive states, EEG-TransNet outperforms other models, especially at longer signal lengths, confirming its ability to generalize well even in more nuanced clinical EEG applications.

Furthermore, the incorporation of the Local Self-Attention Block proved crucial in enhancing the learning of spatial dependencies across EEG channels, leading to improved model performance. The comparison experiments with and without the Local Self-Attention Block clearly indicate that learning regional features in a fully data-driven manner results in superior classification accuracy. By enabling the model to dynamically attend to relevant spatial information, this mechanism strengthens the overall feature extraction process. (3) Despite these successes, challenges remain regarding dataset size and variability. As EEG signals exhibit significant inter-subject variability, future work could explore advanced training strategies that better handle cross-subject generalization. The utilization of techniques such as domain adaptation and transfer learning may prove an effective means of mitigating the adverse effects of limited datasets on model performance.

In addition to these technical aspects, it is crucial to consider the explainability of EEG-TransNet, particularly in sensitive applications like emotion recognition and mental health diagnosis. The ability to interpret the model’s predictions can significantly improve trust and adoption in clinical settings. By employing explainable artificial intelligence (XAI) methods, we can better understand how EEG-TransNet processes and utilizes the input data, providing insights into the decision-making process. This is especially important in the context of mental health, where clinicians need to trust the model's outputs to make informed decisions. Future work could explore integrating XAI techniques such as attention maps, feature importance, or saliency maps to offer transparent interpretations of the model's learned features. This would help bridge the gap between EEG-TransNet’s black-box nature and its application in clinical scenarios, where explainability is critical for ethical, regulatory, and safety reasons.

\section{Related Work}
\subsection{Multi-spectrum feature integration}

The effective fusion of information from various frequency bands in electroencephalogram (EEG) data has been the subject of considerable attention in the literature. Li et al. \cite{li2022motor} introduced a novel approach combining Convolutional Neural Networks (CNNs) and Long Short-Term Memory networks (LSTMs) to extract features from three-band EEG time series, yielding promising results in working memory classification tasks. Their method employed Azimuthal Equidistant Projection (AEP) \cite{hinks1929retro} to project electrode positions onto a two-dimensional plane, preserving the three-dimensional topological relationships crucial for accurate feature representation. CE-stSENet model \cite{8995501}, arguing that traditional CNN architectures often result in information loss during the flattening process of EEG signals. To address this issue, they introduced a regularized Graph Neural Network (GNN) designed to better capture the topological relationships inherent in brain networks. Their results indicated significant improvements in model performance through the application of band fusion techniques across multiple frequency bands. Rahib Abiyev et al.\cite{abiyev2020identification} uses deep learning with convolutional neural networks (CNN) to detect epileptic seizures from EEG signals. A cross-validation technique was applied to improve the system's performance, and comparative results with other machine learning approaches were obtained\cite{sannidhan2023detection,umer2022comprehensive,kodipalli2023computational}. Gaowei Xu et al. \cite{xu2019deep} uses a pre-trained VGG-16 CNN model on ImageNet to classify MI EEG signals. The pre-trained model's parameters are transferred to a target CNN model with a different output layer, and then the target model is fine-tuned on the MI dataset\cite{zhang2025cross,li2025oral}. T-A-MFFNet\cite{peng2023ta}, combines time domain, channel, and spatial features of EEG signals using a multi-feature fusion network. It aims to learn more valuable features for driving fatigue detection by utilizing a time domain network, channel attention network, spatial attention network, and a multi-feature fusion network based on a squeeze network. The PMF-CNN \cite{guo2022multi} proposed in this paper utilizes multi-frequency band EEG signals as input and employs three parallel modules to extract multi-dimensional features from spatial, temporal, and frequency domains. A CNN classification module fuses these features to achieve accurate SSVEP-EEG signal classification.

Despite these advancements, many existing deep learning models heavily rely on Recurrent Neural Networks (RNNs), which excel in capturing temporal information. In contrast, CNNs and GNNs often require more extensive architectures to match RNN performance, which can complicate model design and training\cite{xing2025short}. It is evident that there is a necessity for the development of a more comprehensive and efficacious deep learning model, one that is specifically designed for the modelling of EEG data. This need motivated the development of the MEET model, which seeks to leverage the strengths of multiple architectures while addressing their limitations.

\subsection{CNN on EEG}

Convolutional Neural Networks (CNNs) have become a prominent tool in the analysis of EEG signals, especially for tasks such as classification and feature extraction. HS-CNN\cite{dai2020hs} uses a hybrid-scale CNN architecture with a data augmentation method to improve EEG motor imagery classification accuracy. The hybrid-scale CNN architecture allows for different convolution scales to be used for different subjects, while the data augmentation method helps to address the issue of limited training data. Abhishek Iyer et al.\cite{iyer2023cnn} proposed method involves extracting differential entropy features from EEG signals in five frequency bands and using a hybrid CNN-LSTM model to classify emotions. An ensemble model combining the predictions of individual models is used to improve the overall accuracy. MS-CNN model\cite{roy2022efficient} extracts features from multiple scales of EEG signals in different frequency bands. It integrates user-specific features and employs data augmentation techniques to improve classification accuracy. Chakravarthi et al. \cite{chakravarthi2022eeg}uses a hybrid deep learning algorithm, which combines Convolutional Neural Networks (CNNs), Long Short-Term Memory (LSTM) networks, and ResNet-152. This approach aims to overcome the limitations of previous studies and achieve higher accuracy in analyzing EEG signals for emotion recognition and PTSD assessment. Khademi et al.\cite{khademi2022transfer} proposed three hybrid models combining CNN and LSTM for MI-based BCI. These models used transfer learning with pre-trained CNNs (ResNet-50 and Inception-v3) and data augmentation techniques to improve classification accuracy on limited datasets. Sadiq et al.\cite{sadiq2022exploiting} use pretrained CNN models to extract features from EEG signals. It employs multiscale principal component analysis for denoising and continuous wavelet transform for obtaining time-frequency scalograms. These scalograms are then fed into the pretrained models for feature extraction and classification of motor and mental imagery EEG tasks.Li et al. \cite{li2022eeg} EEG seizure prediction method TGCNN combines CNN and Transformer to extract both local and global features from EEG signals. TGCNN first uses STFT to extract time-frequency features, then feeds these features into an alternating structure of CNN and Transformer to model both local features and long-distance dependencies, and finally predicts seizure occurrence using global average pooling and a fully connected layer.F Mattiol et al.\cite{Mattioli_2021} use a 10-layer 1D-CNN with data augmentation to classify brain states from EEG signals. It also employs a transfer learning approach to customize the model to individual subjects using limited data.

Although CNNs have demonstrated notable success in the analysis of EEG signals, two key limitations remain. Firstly, CNNs are primarily effective in capturing local features but often struggle to model the global spatiotemporal dependencies inherent in EEG data, particularly when dealing with long-range temporal relationships. Secondly, CNN architectures are prone to overfitting when trained on limited datasets, especially in the context of EEG signals, which are characterized by their complexity and variability, leading to reduced generalization performance. These challenges necessitate the integration of other deep learning techniques, such as LSTMs or Transformers, alongside data augmentation strategies to enhance model accuracy and robustness.

\subsection{Transformer on EEG}

With regards to electroencephalography (EEG) analysis, the recent incorporation of Transformer architectures has emerged as a notable research trend. These models exhibit superior capabilities in feature extraction and long-range dependency modeling, enabling them to effectively process time-series data. The attention mechanisms inherent in Transformers allow for the identification of relevant temporal patterns, facilitating enhanced interpretation of EEG signals and improving performance in various applications, including clinical diagnosis and cognitive state assessment.
Yonghao Song et al. \cite{song2022eeg} propose EEG Conformer combines two modules: a convolutional module to capture local temporal features and a self-attention module to capture long-term dependencies within the EEG signals. This allows for more accurate classification of EEG data.Sun et al.\cite{sun2021eeg} developed multiple Transformer-based models for motor imagery (MI) EEG classification, achieving superior performance compared to existing state-of-the-art methods. Our findings indicate that the attention mechanism, combined with CNN features, significantly enhances classification accuracy, particularly by leveraging insights from motor cortex activity and employing positional embedding techniques. Wang et al. \cite{wang2022transformers} present a novel approach to emotion recognition from EEG data based on the Transformer model. Their method employs a hierarchical representation, capturing spatial dependencies from the electrode level to the brain region level. By utilizing transformer encoders, our approach integrates information across different brain regions and emphasizes key areas, demonstrating outstanding performance on the DEAP and MAHNOB-HCI databases, with visualizations highlighting the importance of the pre-frontal, frontal, temporal, and parietal lobes in emotion recognition.
B Abibullaev et al.\cite{abibullaev2023deep} provide a comprehensive survey of transformer applications in BCIs. It discusses the advantages and limitations of transformers, their applications in various BCI domains, and challenges such as computational overhead and interpretability. The paper aims to guide researchers and practitioners in understanding the transformative potential of transformers in BCIs. Song et al.\cite{song2021transformer} propose a novel EEG decoding method that leverages an attention mechanism to enhance relevant spatial features and effectively captures global dependencies by slicing data in the time dimension. J Xie et al.\cite{xie2022transformer}propose a Transformer-based model for motor imagery EEG classification that incorporates positional embedding modules with the aim of improving classification performance. The model attains the highest level of performance currently achievable on the PhysioNet dataset.Y Du et al.\cite{du2022eeg} use a transformer-based methodology is proposed for EEG subject classification, wherein characteristics are derived from both the temporal and spatial domains through the utilisation of a self-attention mechanism. The efficacy of this approach has been evaluated on a range of states, and it has been demonstrated to outperform existing methods without the necessity for manual feature extraction.  Siddhad et al.\cite{siddhad2024efficacy} employ a transformer-based approach for EEG classification, whereby the raw EEG data is classified directly without the need for manual feature extraction. This approach has been shown to achieve state-of-the-art results on both local and public datasets for age, gender, and mental workload classification.

Despite the promising advancements of Transformer-based models in EEG analysis, two notable limitations persist. First, the computational complexity and high resource demands of Transformer architectures make them less practical for real-time EEG applications, particularly when dealing with large datasets or requiring low-latency processing. Second, while Transformers excel in capturing long-range dependencies, they may struggle with interpretability, making it challenging to explain which aspects of the EEG signals are driving model predictions, especially in sensitive applications like clinical diagnosis.

\section{Methods}
\subsection{Overview of Our Network}

This study is a retrospective analysis of three publicly available, de-identified datasets: BETA, SEED, and DEAP EEG. The data were accessed and downloaded for research purposes between October 1, 2025, and October 10, 2025. Throughout the entire process of data collection and analysis, the authors did not have access to any information that could identify individual participants, as all datasets were fully anonymized by their original providers.

\begin{figure}[h]
	\centering
	\includegraphics[width=1.0\textwidth]{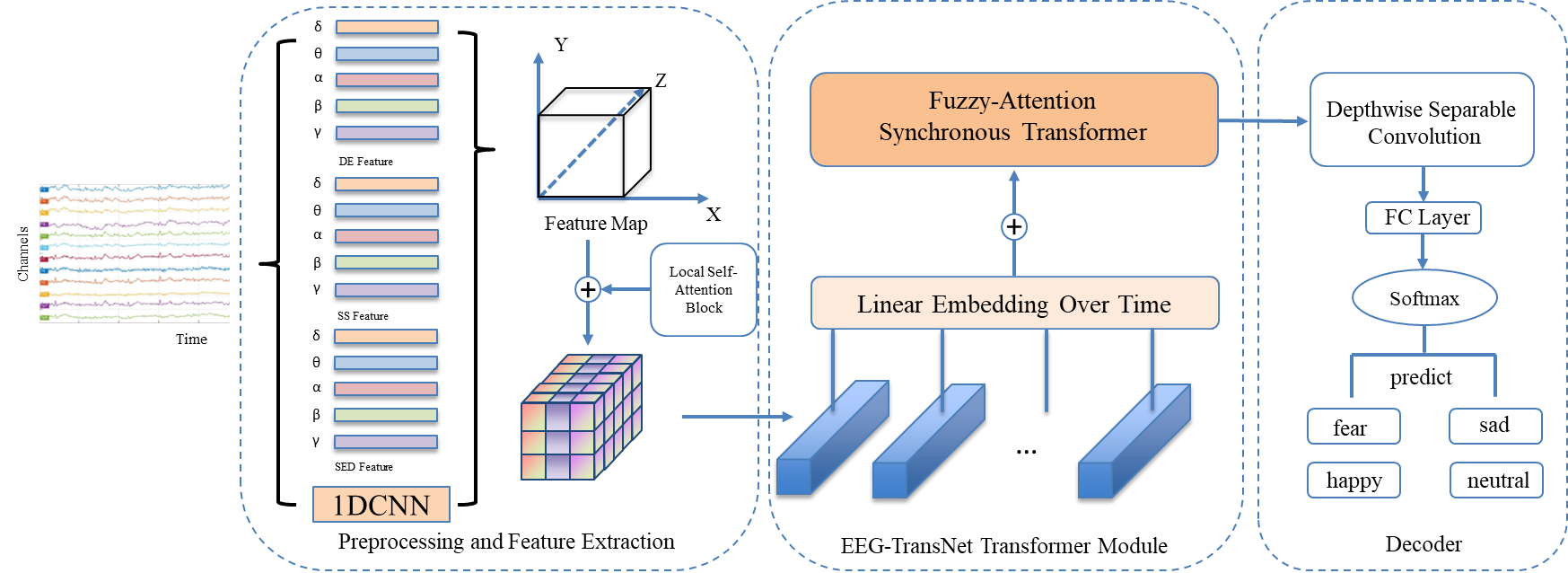}
	\caption{Overall flow chart of the model}
	\label{all}
\end{figure}

This network architecture is designed for processing electroencephalography (EEG) signals, comprising three primary modules that work in tandem. Firstly, the preprocessing and feature extraction module employs a ResNet to extract features across different frequency bands, which are then enhanced through a local self-attention mechanism. Subsequently, the EEG-TransNet Transformer module captures global dependencies in the time series data using a long-cross attention mechanism. Finally, the decoder module further processes these features with depthwise separable convolutions and a fully connected layer, ultimately yielding emotion state predictions—such as fear, sadness, happiness, and neutrality—via a Softmax classifier. Figure \ref{all} is the overall flow chart.

\subsection{Multi-Band Feature Extraction}

To improve the signal-to-noise ratio (SNR) of the input EEG signals, we applied Discrete Wavelet Transform (DWT) for denoising. First, the EEG signal was decomposed into sub-signals across different frequency bands using DWT. To remove noise, we employed a soft thresholding method, where the threshold was set by estimating the standard deviation of the noise in the detail coefficients of the EEG signal. The noise standard deviation, denoted as \( \sigma \), was calculated using the Median Absolute Deviation (MAD) method, as shown by the following equation:

\begin{equation}
    \sigma = \frac{\text{median}(|X_i - \text{median}(X)|)}{0.6745}
\end{equation}

where \( X_i \) represents the coefficients of the detail components. Based on this standard deviation, the threshold \( \lambda \) was computed as:

\begin{equation}
    \lambda = \sigma \sqrt{2 \log n}
\end{equation}

where \( n \) denotes the length of the signal. After applying this threshold, coefficients exceeding the threshold were retained, while those below the threshold were either reduced or set to zero. Finally, the denoised EEG signal was reconstructed via the inverse wavelet transform. Compared to traditional low-pass filtering, the DWT-based denoising method effectively eliminates high-frequency noise from EEG signals while preserving key signal components.

For each EEG signal trial, we segmented the signal into 2-second epochs with a 50\% overlap. This approach enhances temporal resolution, allowing for more precise capture of transient events and dynamic changes within the signal. To compute spectral power (SS) features, we applied the Discrete Fourier Transform (DFT) to each 2-second segment and calculated the magnitude at each frequency point. This overlapping segmentation method makes better use of the data, especially for identifying complex signal changes or detecting shorter events.

The SS feature is computed within each frequency band. The Differential Entropy (DE) feature is extracted using the same five frequency bands as the SS feature. These bands are approximated by a Gaussian distribution to compute the five DE features. Additionally, we introduced Multiscale Entropy (MSE) features to describe the complexity and irregularity of EEG signals across different time scales. MSE, calculated by applying sample entropy on multiple scales of the signal, provides a comprehensive reflection of the dynamic characteristics of EEG signals. The mathematical formulations for the three types of features are as follows:

\begin{equation}
    SS(X) = \int_{f_1}^{f_2} |X(f)| df
\end{equation}

\begin{equation}
    DE(X) = \frac{1}{2} \log(2 \pi e \sigma^2)
\end{equation}

\begin{equation}
    MSE(X) = - \sum P \log P
\end{equation}

where \( X \) denotes the 1-second input EEG signal from a channel, \( X(f) \) represents the frequency domain representation of the signal \( X \), and \( | \cdot | \) denotes the modulus operator. \( f_1 \) and \( f_2 \) correspond to the lower and upper bounds of the signal bandwidth, \( \sigma \) is the standard deviation of the Gaussian distribution, and \( P \) represents the probability distribution of the signal across different scales, while \( \pi \) and \( e \) are constants. By incorporating multiscale entropy, we can more effectively capture the dynamic temporal variations in EEG signals, enhancing classification and recognition capabilities.

Although one-dimensional DE features can be easily fused across multiple frequency bands, this operation overlooks the spatial distribution of electrodes. To address this issue, we employed Lambert Azimuthal Equal-Area Projection, which maps three-dimensional electrode coordinates onto a two-dimensional plane. Lambert projection is commonly used in Geographic Information Systems (GIS) and preserves the true proportionality of areas, making the mapped feature distribution more representative spatially. The one-dimensional DE feature vector is reorganized into a two-dimensional scatter plot. We then used Clough-Tocher interpolation to generate a 32 × 32 map.

To integrate information across different channels and recognize correlations between different time points, we used a 1D-CNN to generate rich three-dimensional feature maps and effectively capture local features in the time series. The EEG segments were detrended and normalized before being fed into the 1D-CNN. The 1D-CNN module utilizes multiple depthwise convolutional operations to efficiently extract features from EEG channels. These convolutional layers apply multiple filters to the input signal, capturing key information across different frequency ranges. In each layer, depthwise convolutions process each channel independently, while sliding filters over the temporal dimension to detect latent time dependencies and local features within the signal. This process ultimately generates a 3D feature map rich in spatial and temporal information, laying the groundwork for subsequent feature analysis and processing. Through this mechanism, the 1D-CNN fully exploits key EEG signal features, improving the model's performance. The result  of the 1D-CNN process is a three-dimensional feature matrix with dimensions \( 32 \times 32 \times B \), where \( B \) represents the feature length of the final layer's output. This 3D feature matrix effectively integrates information extracted from multiple channels, providing a comprehensive foundation for further analysis. For better understanding and visualization, we describe the axes of this feature matrix using a three-dimensional coordinate system. In this coordinate system, the X-axis represents temporal information, reflecting the dynamic changes of the signal over time; the Y-axis represents spatial information, displaying the distribution of features across different EEG channels; and the Z-axis contains convolutional feature information, providing the feature responses of each channel under different convolutional filters. This multi-dimensional representation allows for a comprehensive analysis of the EEG signal’s complexity, enhancing the model's performance in specific tasks.

Finally, we concatenate the SS, DE, MSE, and 1D-CNN-extracted features and input them into the EEG-TransNet encoder, The EEG features are encoded in a method that is consistent across all three categories: regional, temporal, and synchronous. The decoder is tasked with the decoding of these features as well as the inference of outcomes based on the specific task at hand.

\subsection{Local Self-Attention Block}

When processing EEG signals, the complexity of spatiotemporal features and the multi-band nature of the data make it challenging for traditional convolutional neural networks (CNNs) and recurrent neural networks (RNNs) to effectively capture long-range dependencies and multi-scale features. To address this issue, we introduce a Local Self-Attention Block, designed to adaptively focus on important regions of the signal and enhance the modeling of spatiotemporal dependencies. This module combines local self-attention mechanisms with depthwise separable convolutions, allowing the model to efficiently process multi-band EEG data while capturing fine-grained relationships between different frequency bands and channels, all while maintaining computational efficiency.

\begin{figure}[h]
	\centering
	\includegraphics[width=1.0\textwidth]{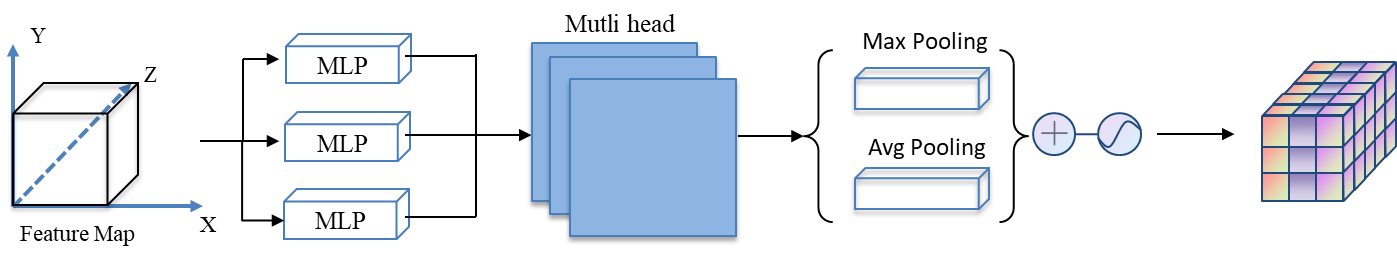}
        \caption{Local Self-Attention Block}
	\label{img_self_attention}
\end{figure}

This module is designed to process three-dimensional multi-band feature tensors. The input feature map is denoted as \( F \in \mathbb{R}^{H \times W \times C} \), where \( H \) and \( W \) represent the height and width of the feature map, respectively, and \( C \) represents the number of bands, as shown in Fig \ref{img_self_attention}.

Initially, the feature map is processed through multiple multilayer perceptrons (MLPs), each of which performs a linear projection on different input bands to generate distinct feature representations. The output of the \( i \)-th MLP is denoted as \( F_i \), and is calculated as:
\[
F_i = \text{MLP}_i(F), \quad i = 1, 2, 3
\]

Subsequently, the generated features are passed through a multi-head attention mechanism. The attention weights are computed via linear projections of the query (\( Q \)), key (\( K \)), and value (\( V \)) matrices. For each band \( i \), the linear projections for the query, key, and value are formulated as:

\begin{equation}
    Q_i = F_i W_Q, \quad K_i = F_i W_K, \quad V_i = F_i W_V
\end{equation}

where \( W_Q \), \( W_K \), and \( W_V \) are the weight matrices for the query, key, and value, respectively.

The attention output for each band is computed based on the standard self-attention formulation:
\begin{equation}
    \text{Attention}(Q_i, K_i, V_i) = \text{softmax} \left( \frac{Q_i K_i^T}{\sqrt{d}} \right) V_i
\end{equation}

After obtaining the multi-head attention output, the feature map undergoes max pooling and average pooling operations, producing global maximum features \( F_{\text{max}} \) and average features \( F_{\text{avg}} \), respectively:

\begin{equation}
    F_{\text{max}} = \text{MaxPool}(F), \quad F_{\text{avg}} = \text{AvgPool}(F)
\end{equation}

The results of the max pooling and average pooling are then element-wise summed to obtain the final feature representation \( F_{\text{fuse}} \):

\begin{equation}
    F_{\text{fuse}} = F_{\text{max}} + F_{\text{avg}}
\end{equation}

Finally, the fused features are processed through a LeakyReLU activation function, given by:
\begin{equation}
    F_{\text{out}} = \text{LeakyReLU}(F_{\text{fuse}})
\end{equation}

The resulting output feature tensor \( F_{\text{out}} \in \mathbb{R}^{H \times W \times C} \) contains rich contextual information across different bands, and the LeakyReLU activation enhances the ability to capture dependencies among the various bands.

Local Self-Attention Block, with its powerful spatiotemporal modeling capabilities and flexible feature weighting mechanism, significantly enhances EEG signal analysis performance. By leveraging multi-head attention mechanisms, the module captures dynamic feature dependencies across multiple scales and bands, improving the accuracy of tasks such as emotion recognition and brain activity classification. The introduction of this module not only optimizes EEG decoding effectiveness but also provides computational efficiency benefits for real-time applications and large-scale data processing.

\subsection{Fuzzy-Attention Synchronous Transformer}

When processing EEG signals, noise and uncertainty pose significant challenges that can degrade model performance. Traditional neural network methods, such as Convolutional Neural Networks (CNNs) and Recurrent Neural Networks (RNNs), while effective for sequential data, often struggle with the complexity and high noise environments inherent in EEG signals. To overcome this, we introduce the Fuzzy-Attention Synchronous Transformer (FAST) module, which integrates fuzzy logic with the self-attention mechanism of Transformers. The FAST module captures spatiotemporal dependencies in EEG data while effectively managing uncertainty and noise. By incorporating fuzzy membership functions, the module adapts to the dynamic fluctuations of EEG signals, enhancing robustness and performance in tasks like emotion recognition and brain activity classification.

The FAST module models the spatiotemporal dependencies of input features, capturing interactions across different channels or features while handling noise and uncertainty through differentiable fuzzy logic. The input to the FAST module is the output of the Local Self-Attention Block, denoted as \( z_4 \in \mathbb{R}^{H \times W \times C} \). This 3D matrix \( z_4 \) is first divided along the convolutional feature dimension into \( C \) 2D submatrices, where each submatrix is represented as \( X_{\text{fuzzy},i} \in \mathbb{R}^{H \times W} \) (for \( i = 1, 2, 3, \dots, C \)), corresponding to features from different channels.

For each submatrix, the vector \( X_{\text{fuzzy},i} \) is sequentially extracted in terms of the spatial dimensions and passed through a linear module. The vector \( X_{\text{fuzzy},(i,s)} \) is defined as a patch in the FAST module and is linearly mapped into a latent vector \( z_{\text{fuzzy},(0),(i,s)} \) using a learnable matrix \( M \), as described by the following equation:

\begin{equation}
    z_{\text{fuzzy},(0),(i,s)} = M X_{\text{fuzzy},(i,s)} + e_{\text{pos},(i,s)},
\end{equation}

where \( e_{\text{pos},(i,s)} \) represents the positional embedding that codifies the positional nuances of each EEG signal as it fluctuates throughout the designated time frame.

The FAST module incorporates differentiable fuzzy logic within the attention mechanism to enhance its ability to handle uncertainty and noise. Unlike traditional multi-head self-attention mechanisms, where the similarity between the query (Q) and key (K) is computed via dot product, in the FAST module, the query \( Q \) and key \( K \) are first fuzzified using membership functions. These membership functions transform the inputs into fuzzy values, represented as follows:

\begin{equation}
    \mu(x) = \max\left(0, \min\left(\frac{x - a}{b - a}, 1, \frac{c - x}{c - b}\right)\right),
\end{equation}

where \( a \), \( b \), and \( c \) are the parameters of the membership function, which define the fuzzy set range. These parameters are learned during training and are critical for controlling the fuzziness of the query and key. Specifically, \( a \) and \( c \) represent the lower and upper bounds of the fuzzy set, and \( b \) defines the transition point where the membership function reaches its peak value of 1. This setup allows for adaptive fuzzification that can be optimized through gradient descent, ensuring that the module remains differentiable and suitable for end-to-end learning. The fuzzified query and key are denoted as \( \mu(Q) \) and \( \mu(K) \), respectively.

For each input token represented by \( z_{\text{fuzzy},(0),(i,s)} \), the fuzzy self-attention mechanism in the FAST module computes the fuzzified query \( \mu(q_{\text{fuzzy},(l,a),(i,s)}) \), fuzzified key \( \mu(k_{\text{fuzzy},(l,a),(i,s)}) \), and value \( v_{\text{fuzzy},(l,a),(i,s)} \), as given by the following equations:

\begin{equation}
    q_{\text{fuzzy},(l,a),(i,s)} = W'^{(l,a)}_Q \text{LN}(z_{\text{fuzzy},(l-1),(i,s)}) \in \mathbb{R}^{D_h},
\end{equation}

\begin{equation}
    k_{\text{fuzzy},(l,a),(i,s)} = W'^{(l,a)}_K \text{LN}(z_{\text{fuzzy},(l-1),(i,s)}) \in \mathbb{R}^{D_h},
\end{equation}

\begin{equation}
    v_{\text{fuzzy},(l,a),(i,s)} = W'^{(l,a)}_V \text{LN}(z_{\text{fuzzy},(l-1),(i,s)}) \in \mathbb{R}^{D_h}.
\end{equation}

Here, \( W'^{(l,a)}_Q \), \( W'^{(l,a)}_K \), and \( W'^{(l,a)}_V \) represent the weight matrices for query, key, and value, respectively, and \( \text{LN} \) denotes layer normalization. The fuzzy synchronous self-attention (FSSA) scores \( \alpha_{\text{fuzzy},(l,a),(i,s)} \), computed using fuzzy logic and defuzzification, are given by:

\begin{equation}
    \alpha_{\text{fuzzy},(l,a),(i,s)} = \sigma \left( \mu(Q) \land \mu(K) \right) = \sigma \left( \frac{\mu(q_{\text{fuzzy},(l,a),(i,s)})}{\sqrt{D_h}} \cdot \left[ \mu(k_{\text{fuzzy},(l,a),(0,0)}), \{\mu(k_{\text{fuzzy},(l,a),(i,s)}) \}_{s = 1,...,S} \right] \right) \in \mathbb{R}^S,
\end{equation}

where \( \land \) denotes the fuzzy logical "AND" operation, and \( \sigma \) represents the softmax activation function that computes the final fuzzified attention scores. This approach allows the FAST module to handle spatiotemporal features with high uncertainty more flexibly, enhancing the robustness of the model to noisy data.

The output features \( z_{\text{fuzzy},(l),(i)} \) from the encoding block are obtained by combining the vectors from all processing stages, and the final output of the FAST module is represented as \( z \in \mathbb{R}^{C \times H \times W} \).

By integrating fuzzy logic with Transformer architecture, the FAST module effectively addresses the noise, uncertainty, and complex spatiotemporal dependencies inherent in EEG signals. The module demonstrates outstanding performance in emotion recognition and brain activity classification tasks, providing greater robustness and flexibility in highly uncertain environments. By fuzzifying the query and key components, the FAST module not only improves model performance but also enhances adaptability in dynamic data environments, offering a new and effective tool for EEG signal analysis.

\subsection{Decoder}

The objective of EEG-TransNet is to facilitate the uniform capture of temporal, regional, and synchronous features for the analysis of brain activities based on EEG data. In contrast to conventional Transformer decoders, which depend on multi head self-attention mechanisms to decode the outputs, EEG-TransNet utilises a network architecture based on depthwise separable convolutions to achieve task-specific processing. This architectural approach facilitates the efficient decoding of multi-dimensional EEG features while reducing computational complexity. This technique comprises two distinct operations, namely depthwise convolution and pointwise convolution. The former performs spatial convolutions independently over each input channel, whereas the latter applies a 1×1 convolution across channels to combine the features. By employing this methodology, the total number of parameters and the computational cost are significantly reduced. Consequently, training can be expedited, and the memory required can be diminished.

In the first layer, the standard convolution is replaced by a depthwise separable convolution, as described by the following formula:

\begin{equation}
    X_1 = \text{DepthwiseConv}(X, w_1^{\text{depth}}) + \text{PointwiseConv}(X_1, w_1^{\text{point}})
\end{equation}

where \( w_1^{\text{depth}} \) represents the weights of the depthwise convolution, and \( w_1^{\text{point}} \) represents the weights of the pointwise convolution. Similarly, in the second layer, depthwise separable convolution is applied as follows:

\begin{equation}
    X_2 = \text{DepthwiseConv}(X_1, w_2^{\text{depth}}) + \text{PointwiseConv}(X_2, w_2^{\text{point}})
\end{equation}

This decomposition significantly reduces the number of parameters and enhances the model’s computational efficiency.

In addition to the convolutional improvements, we incorporate a hybrid L1-L2 regularization strategy to better control model complexity. L1 regularization encourages sparsity in the model by pushing certain weights towards zero, promoting feature selection, while L2 regularization prevents the excessive growth of weight values, thereby mitigating overfitting. The combined loss function with this hybrid regularization is formulated as:

\begin{equation}
    \text{Loss} = \frac{1}{D_n} \sum_{i=1}^{D_n} -\log \left( p_i (y_i) \right) + \lambda_1 \sum |w| + \lambda_2 \sum w^2
\end{equation}

where \( \lambda_1 \) and \( \lambda_2 \) are the coefficients for L1 and L2 regularization, respectively, and \( w \) represents the weights of the network. This hybrid regularization approach helps maintain the sparsity of the model while simultaneously constraining weight magnitudes, ultimately enhancing generalization and reducing overfitting tendencies.

\section{Experiment}

\subsection{Public Data Set}
To assess the performance of our proposed methods, we conducted experiments on three publicly available EEG datasets.

The BETA dataset includes data collected from 70 healthy participants. Each participant underwent 4 experiments, with 40 trials conducted per experiment. The EEG signals were captured using 64 channels, and the data were down-sampled at a rate of 250 Hz. The labels consist of 40 harmonic frequency categories in the range of 8 to 15.8 Hz, with intervals of 0.2 Hz. The time length per trial is 2 or 3 seconds.

The SEED dataset comprises EEG recordings from 15 healthy participants. Each participant participated in 3 experiments, with 15 trials conducted per experiment. The EEG data were captured using 62 channels and down-sampled at a rate of 200 Hz. The labels represent three emotional states: positive, neutral, and negative. Each trial lasts for 305 seconds.

The DepEEG dataset involves data collected from 35 participants, including 12 healthy subjects and 23 individuals diagnosed with depression. Each participant underwent 1 experiment with 1 trial. The EEG signals were captured using 6 channels, with a down-sampling rate of 500 Hz. The labels include two categories: depressive and normal control. The time length per trial is at least 480 seconds.

\subsection{Experimental Details}
In this paper, 3 datasets are selected for training, and the training process is as follows:

Step1:Dataset Data Preprocessing

The three datasets underwent a series of preprocessing steps to ensure data quality and consistency. First, ocular artifacts were removed using independent component analysis (ICA). Next, the data were segmented into epochs of appropriate length based on the specific dataset. For the BETA dataset, epochs of 2-3 seconds were extracted, aligning with the label intervals. For the SEED dataset, epochs of 1-2 seconds were selected, considering the shorter time intervals of interest. The DepEEG dataset, with longer trials, allowed for epochs of 1-2 seconds to be extracted. To reduce dimensionality and computational complexity, feature extraction was performed. Time-domain features, such as mean, variance, and standard deviation, were calculated for each epoch. Additionally, frequency-domain features, including power spectral density and wavelet coefficients, were extracted to capture the spectral characteristics of the EEG signals. Finally, the extracted features were normalized to a common range to ensure consistent scaling across different datasets and channels.

Step2:Model Training


To ensure a fair comparison, we evaluated EEG-TransNet alongside several state-of-the-art models that have demonstrated strong performance on publicly available datasets. For consistency, we applied the same testing protocol across all methods. The training batch size for EEG-TransNet and other models was set to 128. EEG-TransNet was optimized using the AdamW optimizer, which incorporates L2 regularization and is better suited for Transformer-based architectures. The learning rate was initialized at 0.001 and decayed using the Step Decay strategy to gradually reduce the learning rate. Each Transformer module contained three encoding blocks, and Dropout layers were appropriately introduced to mitigate overfitting. Early stopping was employed as a training strategy. It is worth noting that all hyperparameters were fine-tuned using the validation set. The models were implemented in Pytorch and trained on an NVIDIA RTX 8000 GPU.

Step3:Indicator Comparison Experiment

Compare the performance metrics of EEG-TransNet with baseline models across various datasets to evaluate classification accuracy, specificity, and sensitivity. Discuss how different combinations of features and model architectures impact the performance metrics, highlighting the strengths of EEG-TransNet in capturing complex EEG dynamics. Analyze the effectiveness of incorporating the Local Self-Attention Block in improving the model's ability to learn spatial dependencies and its impact on overall classification accuracy. Evaluate the results in the context of dataset characteristics, such as signal length and channel count, to determine their influence on model performance. Evaluate the performance of each model by analyzing metrics such as the mean classification accuracy (ACC), sensitivity (SEN), and specificity (SPE), along with their associated standard deviations (SD).

Step4:Experimental Results Analysis

Analyze the experimental results to identify the strengths and weaknesses of EEG-TransNet and baseline models in classifying EEG signals. Discuss the role of temporal, regional, and synchronous features in enhancing model performance, and assess the impact of increasing EEG channel count on accuracy, specificity, and sensitivity. Evaluate the effectiveness of transfer learning in improving the generalization ability of the models across different datasets, particularly in challenging scenarios like depressive EEG signals. Examine the contribution of the Local Self-Attention Block to the model's performance and its implications for future improvements in EEG-based emotion and mental state recognition tasks. 

Step5:Conclusion and Discussion

Summarize the experimental results, highlighting the superior performance of EEG-TransNet in accurately classifying EEG signals across different datasets. Discuss the key advantages of integrating temporal, regional, and synchronous features within a unified architecture, as well as the impact of feature combinations and the Local Self-Attention Block on model effectiveness. Address the limitations related to dataset size and inter-subject variability, and suggest future research directions, including advanced training strategies and the exploration of explainable AI techniques, to further enhance the model's applicability and robustness in clinical and real-world scenarios.

1. Average Classification Accuracy (ACC):

\begin{equation}
\text{ACC} = \frac{TP + TN}{TP + TN + FP + FN}
\end{equation}

Where , \( TP \) refers to the correctly identified positive instances, while \( TN \) denotes the correctly identified negative instances. Conversely, \( FP \) and \( FN \) represent the misclassified instances, with \( FP \) indicating the number of negative instances incorrectly classified as positive, and \( FN \) indicating the positive instances incorrectly classified as negative. These metrics collectively inform the accuracy calculation by measuring the model's ability to distinguish between classes.

2. Sensitivity (SEN):

\begin{equation}
\text{SEN} = \frac{TP}{TP + FN}
\end{equation}

Where \( TP \) is the number of true positives and \( FN \) is the number of false negatives.

3. Specificity (SPE):

\begin{equation}
\text{SPE} = \frac{TN}{TN + FP}
\end{equation}

Where \( TN \) represents the number of true negatives and \( FP \) denotes the number of false positives.

4. Standard Deviation (SD):

\begin{equation}
\text{SD} = \sqrt{\frac{1}{n - 1} \sum_{i=1}^{n} (x_i - \bar{x})^2}
\end{equation}

Where \( n \) represents the overall number of data points, \( x_i \) is the individual metric score for subject  \( i \), and \( \bar{x} \) represents the mean value of the metric across all observations.

These metrics assess the performance of the classification model, with ACC indicating the overall accuracy, SEN measuring the ability to correctly identify positive cases, SPE evaluating the ability to correctly identify negative cases, and SD representing the dispersion of these metrics across different samples.

Algorithm \ref{algorithm1} provides a detailed representation of the training process for the EEG-TransNet network discussed in this paper. It outlines each step from data preprocessing, feature extraction, and network forward pass, to loss calculation and model optimization. This algorithm captures the essential flow of training, ensuring the network effectively learns to recognize emotion states from EEG signals.

\begin{algorithm}[H]
  \SetAlgoLined
  \caption{Training process for the EEG-TransNet network}
  \label{algorithm}
  
  \KwIn{Training dataset: Preprocessed EEG signals with corresponding emotion labels}
  
  \KwOut{Trained EEG-TransNet network}
  
  Initialize EEG-TransNet network architecture\;
  Initialize hyperparameters (learning rate, batch size, etc.)\;
  Initialize the DWT parameters for denoising\;
  Initialize loss function (e.g., categorical cross-entropy)\;
  
  \While{not converged}{
    Randomly sample a batch of training examples from the dataset\;
    Preprocess the data (DWT denoising, segmenting into epochs, etc.)\;
    
    Compute SS, DE, and MSE features from the segmented EEG signals\;
    Project the DE features using Lambert Azimuthal Equal-Area Projection\;
    
    Forward pass through the 1D-CNN to extract local features\;
    
    Pass the extracted features through the Local Self-Attention Block\;
    Apply multi-head attention and pooling operations\;
    
    Forward pass through the Fuzzy-Attention Synchronous Transformer (FAST) module\;
    Perform fuzzy logic operations to handle uncertainty in the data\;
    
    Pass the features through the depthwise separable convolutional layers in the decoder\;
    Apply hybrid L1-L2 regularization to the network parameters\;
    
    Compute the predicted emotion state using a Softmax classifier\;
    Compute the loss between the predicted and true emotion states\;
    
    Calculate evaluation metrics (accuracy, precision, recall, F1 score)\;
    
    Backpropagate the gradients and update network parameters\;
  }
  
  \Return{Trained EEG-TransNet network}
  \label{algorithm1}
\end{algorithm}

\section{Author Contributions}

Xinglong Cui is responsible for the overall design and architecture of the EEG-TransNet model, including the development of the multi-module network structure and feature extraction strategies. He applied Discrete Wavelet Transform (DWT) during the signal preprocessing and multi-band feature extraction process to enhance the signal-to-noise ratio (SNR) of the input EEG signals, ensuring effective feature extraction. Additionally, Xinglong Cui led the design of the Local Self-Attention Block and the Fuzzy-Attention Synchronous Transformer (FAST) module to optimize the modeling of spatiotemporal dependencies in the features.

Dian Gu primarily handles the experimental design and evaluation of the model, implementing the construction of the depthwise separable convolution decoder and conducting model training and validation. He focuses on optimizing the computational efficiency and performance of the model by introducing a hybrid L1-L2 regularization strategy to control model complexity, thereby enhancing its performance in emotion recognition tasks. Furthermore, Dian Gu is responsible for writing the detailed descriptions of the methods section, ensuring the transparency and reproducibility of the research.


\section{Availability of Data and Materials}

The datasets used in this study, including the BETA dataset, SEED dataset, and DepEEG dataset, are publicly available. The BETA dataset can be accessed at \url{https://figshare.com/articles/dataset/The_BETA_database/12264401}, the SEED dataset is available at \url{https://bcmi.sjtu.edu.cn/home/seed/}, and the Deap EEG dataset can be obtained from \url{https://github.com/pratyakshajha/emotion-recognition-by-deap-dataset}. These datasets were utilized to evaluate the models and can be freely accessed for further research purposes.

\section{Funding}
None.

\bibliography{sample}

\end{document}